\documentclass[conference]{IEEEtran}  % Comment this line out if you need a4paper

\usepackage{multicol}
\usepackage{lmodern}
\usepackage[dvipsnames,table,xcdraw]{xcolor}
\definecolor{DodgerBlue}{RGB}{30,144,255}
\usepackage[numbers]{natbib}

\usepackage{hyperref}
\hypersetup{breaklinks=true,bookmarks=true,colorlinks=true,linkcolor=DodgerBlue,urlcolor=DodgerBlue, citecolor=DodgerBlue,}
\usepackage{graphics} % for pdf, bitmapped graphics files
\usepackage{epsfig} % for postscript graphics files
\usepackage{times} % assumes new font selection scheme installed
\usepackage{amsmath} % assumes amsmath package installed
\usepackage{amssymb}  % assumes amsmath package installed
\usepackage[utf8]{inputenc}
\usepackage[ruled,vlined, linesnumbered]{algorithm2e}
\usepackage[margin=1in]{geometry}
\usepackage{subcaption}
\usepackage{booktabs}
\usepackage{multirow}
\usepackage{cleveref}
\usepackage{array}
\usepackage[misc]{ifsym}
\newcommand{\SE}{\mathrm{SE}}
\newcommand{\SO}{\mathrm{SO}}

\title{\LARGE \bf
Push-Grasp Policy Learning Using Equivariant Models and Grasp Score Optimization
}

\author{
Boce Hu\textsuperscript{$\star$1}, 
Heng Tian\textsuperscript{$\star$1}, 
Dian Wang\textsuperscript{\dag 1}, 
Haojie Huang\textsuperscript{1},
Xupeng Zhu\textsuperscript{1},
Robin Walters\textsuperscript{1,2}, 
Robert Platt\textsuperscript{1,2}\\
\textsuperscript{1}Northeastern University \quad\quad
\textsuperscript{2}Robotics and AI
Institute\\
\textsuperscript{$\star$} Equal Contribution \quad \textsuperscript{\dag} Corresponding Author
\\ 
\small \texttt{\{hu.boce; tian.hen; wang.dian; huang.haoj; zhu.xup; r.walters; r.platt\}@northeastern.edu }
}

\begin{document}

\twocolumn[{
\renewcommand\twocolumn[1][]{#1}
\maketitle
    \vspace{-7mm}
\begin{center}
    % \captionsetup{font=footnotesize}
    \includegraphics[width=1\textwidth]{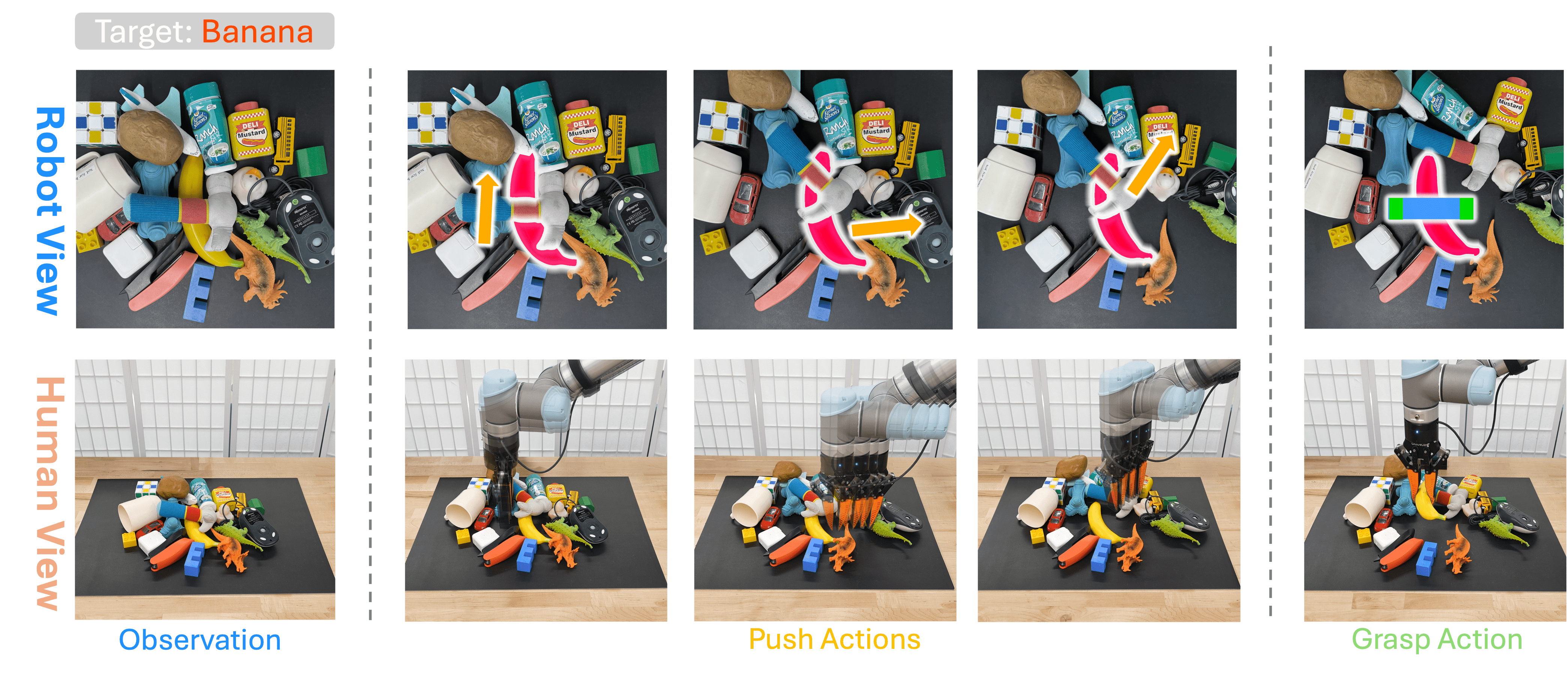}
    \vspace{-0.3cm}
    \captionof{figure}{\textbf{Illustration of the Push-Grasp Workflow.} The target object, specified by human instruction, is highlighted with a red mask (e.g., a banana). At each step, the push action direction is represented by an arrow. Our method iteratively predicts and executes push actions to create sufficient space for grasping the target. The final grasp pose is shown as a blue rectangle, with green blocks indicating the gripper's fingers.
    } 
    \label{fig:teaser}
    \vspace{-0.1cm}
\end{center}
}]

\begin{abstract}

Goal-conditioned robotic grasping in cluttered environments remains a challenging problem due to occlusions caused by surrounding objects, which prevent direct access to the target object. A promising solution to mitigate this issue is combining pushing and grasping policies, enabling active rearrangement of the scene to facilitate target retrieval.
However, existing methods often overlook the rich geometric structures inherent in such tasks, thus limiting their effectiveness in complex, heavily cluttered scenarios. To address this, we propose the Equivariant Push-Grasp Network, a novel framework for joint pushing and grasping policy learning. Our contributions are twofold: (1) leveraging $\SE(2)$-equivariance to improve both pushing and grasping performance and (2) a grasp score optimization-based training strategy that simplifies the joint learning process. Experimental results show that our method improves grasp success rates by 49\% in simulation and by 35\% in real-world scenarios compared to strong baselines, representing a significant advancement in push-grasp policy learning.

\end{abstract}

\section{Introduction}
Effective grasping of target objects in cluttered environments is crucial for many robotic manipulation tasks. Recent grasp learning methods \cite{lim2024equigraspflow, breyer2021volumetricgraspingnetworkrealtime, huang2022edgegraspnetworkgraphbased, hu2024orbitgrasp} have achieved promising performance but typically focus on lightly cluttered scenes or decluttering tasks, where targets are not heavily occluded. Grasping target objects in densely cluttered scenes remains challenging due to severe occlusion and limit space for gripper fingers. Recent research explores the synergy between non-prehensile (pushing) and prehensile (grasping) actions to enhance grasping performance in such scenarios~\cite{zeng2018learning, Tang2021, xu2021efficient, wang2022self}. Nevertheless, current push-grasp frameworks for goal-conditioned object retrieval still have several limitations. Firstly, conventional network architectures struggle to represent the extensive state and action spaces associated with the push-grasp task, leading to poor generalization in novel, cluttered scenarios. Second, these methods are often sample-inefficient, as they require extensive data, heavy augmentation, and long training times~\cite{zeng2018learning}. Lastly, many existing approaches involve complex training processes, often relying on alternating optimization between grasping and pushing prediction networks\cite{xu2021efficient,wang2022self}.

In this paper, we introduce the \textbf{Equivariant PushGrasp (EPG)} Network, a novel framework for efficient goal-conditioned push-grasp policy learning in cluttered environments. EPG leverages inherent task symmetries to improve both sample efficiency and performance. Specifically, we model the pushing and grasping policies using $\SE(2)$-equivariant neural networks, embedding rotational and translational symmetry as an inductive bias. This design substantially enhances the model's generalization and data efficiency. Furthermore, we propose a self-supervised training approach that optimizes the pushing policy with a reward signal defined as the change in grasping scores before and after each push. This formulation simplifies the training procedure and naturally couples the learning of pushing and grasping.

In summary, our contributions are threefold. First, we propose a fully-$\SE(2)$-equivariant push-grasp framework that leverages the symmetry of environment dynamics as an inductive bias to boost policy learning efficiency. Second, we introduce a novel training strategy that treats the learned grasping policy as part of the environment, serving as a critic to guide and optimize the learning of the pushing policy. Lastly, extensive experiments in both simulation and real-world environments validate the effectiveness of our approach. Our proposed EPG achieves a \textbf{49\%} improvement in grasp success rates in simulation and a \textbf{35\%} improvement in real-world scenarios compared to prior baselines~\cite{xu2021efficient,wang2022self}.

%%%%%%%%%%%%%%%%%%%%%%%%%%%%%%%%%%%%%%%%%%%%%%%%%%%%%%%%%%%%%%%%%%%%%%%%%%%%%%%%

\section{Related Work}

\subsection{Pushing and Grasping in Cluttered Environments}
Target grasping in cluttered environments is challenging due to object overlap, occlusions, and the need for precise selection in densely populated scenes. Early approaches~\cite{mahler2017dex,morrison2018closing} evaluated $\SE(2)$ grasp configurations from top-down images but primarily focused on isolated objects or sparse environments. Recent advances~\cite{zhou2018fully,AntipodalRoboticGrasping, satish2019policy,zhu2022sample} have made progress toward handling denser scenes, but often struggle in highly cluttered environments or when specific target objects must be retrieved.

Non-prehensile manipulations, such as pushing, provide effective solutions for separating objects or clearing clutter. The synergy between pushing and grasping has been widely studied to explore their combined potential. \citet{zeng2018learning} established a self-supervised framework for unified push-grasp policies using deep Q-learning, demonstrating the benefit of strategic pushing in creating grasp opportunities, but with limited generalization to complex environments. ~\citet{Tang2021} extended the action space from $\SE(2)$ to $\SE(3)$ to enable more flexible and precise 6-DoF grasping. Building on~\cite{zeng2018learning}, \citet{xu2021efficient} and \citet{wang2022self} proposed goal-conditioned push-grasp strategies for targeted retrieval. However, these methods suffer from simplistic network architectures and complex training procedures which limit their effectiveness in highly dynamic and cluttered environments. Compared with these methods, our approach incorporates $\SE(2)$-equivariance to enhance the representational capacity of both pushing and grasping policies. We also introduce a simplified and straightforward training pipeline, which reduces the training complexity and hyperparameter sensitivity, thereby improving the generalizability and robustness.

\subsection{Equivariance in Robot Learning}
The integration of symmetries and equivariance properties into robotic policy learning has been proven to enhance both efficiency and performance~\cite{wangsurprising, huang2022equivariant, simeonov2022neural, eisner2024deep, ryu2022equivariant, wang2024general, tie2024seed}. In deep reinforcement learning (DRL), recent methods~\cite{wang2022mathrm, wang2022equivariant, zhu2022sample, nguyen2023equivariant} demonstrate remarkable improvements in performance and convergence speed for $\SE(2)$ manipulation tasks. Similarly, equivariance has also shown effective in imitation learning (IL) ~\cite{huang2022equivariant, hu2024orbitgrasp, huang2024fourier, wang2024equivariant, gao2024riemann}. Closest to our approach are \cite{zhu2022sample, wang2022equivariant}, which establish foundational techniques for $\SE(2)$-equivariant policy learning. Unlike these prior methods that directly train a single equivariant policy via IL or RL to accomplish the entire task, our method introduces a novel pipeline that first employs IL to train a grasping network, which subsequently serves as the environment for DRL-based training of a pushing network. This two-step training strategy improves both training efficiency and generalization capabilities.

%%%%%%%%%%%%%%%%%%%%%%%%%%%%%%%%%%%%%%%%%%%%%%%%%%%%%%%%%%%%%%%%%%%%%%%%%%%%%%%%

\section{Method}
\begin{figure*}[!ht]
\centering
\includegraphics[width=1\linewidth]{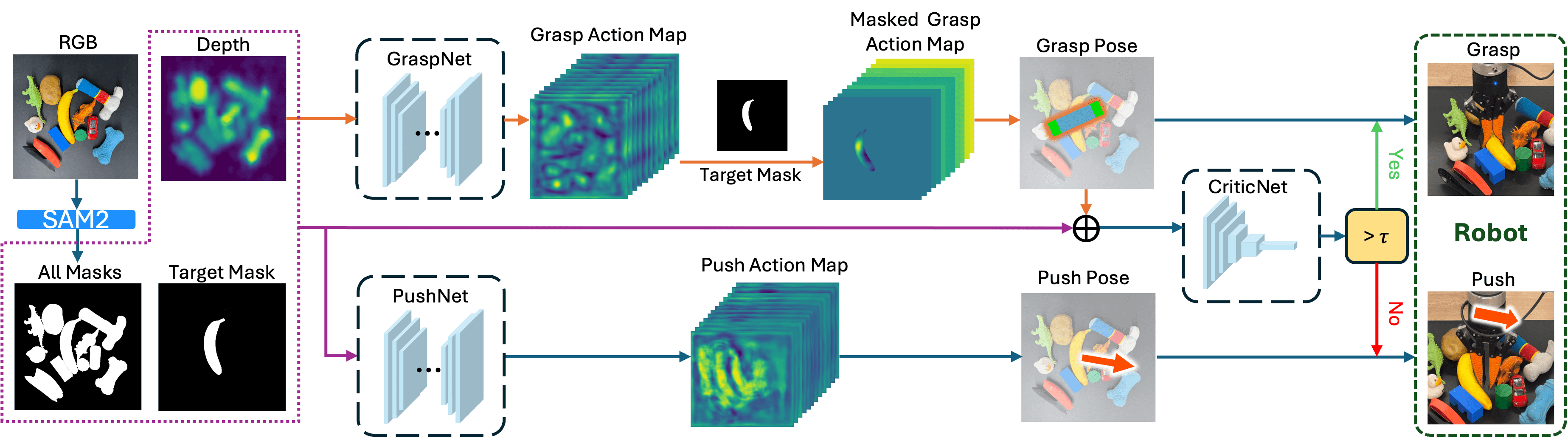}
\caption{
Given an RGB-D observation, SAM2\cite{ravi2024sam2} generates a set of object masks. GraspNet and PushNet then use the depth image and these masks to predict candidate grasp and push actions. The target object's grasp pose is filtered using its corresponding mask, and the best candidate is selected. Finally, CriticNet evaluates the selected grasp pose against a threshold $\tau$ to determine whether to execute the grasp or a push action.
}
\label{fig:pipeline}
\vspace{-0.5cm}
\end{figure*}

\subsection{Problem Statement}
The target object retrieval task in cluttered environments requires the agent to execute a series of push actions to clear obstructions, followed by a final grasping action to pick up the target. At each time step $t$, the agent observes the state $O_t\in \mathcal{O}$ and the specified target object, represented by its mask $k\in \mathcal{K}$, where $\mathcal{O}$ denotes the observation space and $\mathcal{K}$ is the set of all object masks in the scene. We use a top-down RGB-D image as the observation, i.e., $O_t \in \mathbb{R}^{4 \times h \times w}$.  The agent then selects an action $a_t \in \mathcal{A}$, where $\mathcal{A} = \mathcal{A}{push} \cup \mathcal{A}{grasp}$ includes all top-down grasps and horizontal pushes. Each action is represented as a tuple $(\textit{type}, \textit{pose})$, with $\textit{type} \in \{push, grasp\}$ and $pose \in \SE(2)$. To model the policy, we represent the end-effector pose as a distribution over discretized $\SE(2)$ actions, encoded as a pixel-wise dense action map of shape $n \times h\times w$. Here, the spatial translation component is discretized into $h \times w$ bins and the rotation component into $n$ bins, where each pixel in the action map corresponds to a translation and each channel to a rotation angle, so the entire map defines a function over the discretized $\SE(2)$ space, similar to prior works~\cite{wang2022robot,wang2022mathrm, zhu2022sample}.

\subsection{Overview of the Approach}
The key contribution of our work is a novel push-grasp framework for efficient target object retrieval. As illustrated in Figure~\ref{fig:pipeline}, our workflow consists of three key components: a CriticNet $\sigma$, a GraspNet $\pi$, and a PushNet $\phi$.  At each time step, GraspNet and PushNet generate a grasp action and a push action with respect to the target object. CriticNet then evaluates the grasp action by assigning it a score. If the score exceeds a predefined threshold $\tau$ or the maximum number of push attempts is reached, the grasp action is executed. Otherwise, the push action is executed, and the process repeats with an updated observation. In the following subsections, we first describe the training process for each agent, followed by the design of equivariant networks. 

\subsection{Two-Step Agent Learning}
Previous works often rely on complex alternating training procedures, where grasp and push networks are optimized iteratively. This process can lead to unstable convergence and difficulty in balancing the learning dynamics between different networks. In contrast, we propose a simple two-step training process. First we train a universal, goal-agnostic GraspNet together with a CriticNet that evaluates predicted grasps and returns a score. Once they are trained, we use the difference in grasp scores before and after pushing, computed from the CriticNet, as a reward signal to train a goal-conditioned PushNet. This decoupled training strategy eliminates the need for alternating optimization and the associated scheduling-related hyperparameters, making the training procedure more stable, controllable, and efficient.

\textbf{Step 1: GraspNet and CriticNet Training.} 
We first train a universal target-agnostic GraspNet $\pi$ and a target-conditioned CriticNet $\sigma$ using supervised learning. 
We collect a grasping dataset in simulation containing each step observation $O_t$, object mask sets $\mathcal{K}$, grasp poses, and binary success labels. GraspNet $\pi$ takes only the depth channel $D_t\in \mathbb{R}^{h\times w}$ from $O_t$ as input and outputs dense, pixel-wise grasp score maps for all objects in the entire scene, i.e., $\pi: \mathbb{R}^{ h\times w} \to \mathbb{R}^{n\times h\times w}$. Each entry in the Q-map represents a predicted grasp score, indicating the grasp quality at a specific location and orientation, where $h \times w$ corresponds to the spatial resolution and $n$ denotes the number of grasp orientations considered.

The model $\pi$ is trained using Binary Cross Entropy (BCE) loss, defined as: 
\begin{equation}
    \mathcal{L}_{\pi} = - \sum_{a} \left[ y_{a} \log Q_{a} + (1 - y_{a}) \log (1 - Q_{a}) \right]
    \vspace{-0.1cm}
\end{equation}
where $Q_{a}$ is the predicted score for grasp pose $a$, and $y_{a}$ is the label of whether the grasp is successful. Since simulation allows us to collect a large number of grasp poses for each mask in each $O_t$, multiple grasp strategies can be generated for a single scene. Due to the independent, pixel-wise optimization of each grasp pose in the above formulation, the network can capture the multi-modality of grasping by learning diverse grasp strategies in each $O_t$.

\begin{figure*}[!t]
\centering
\includegraphics[width=0.9\linewidth]{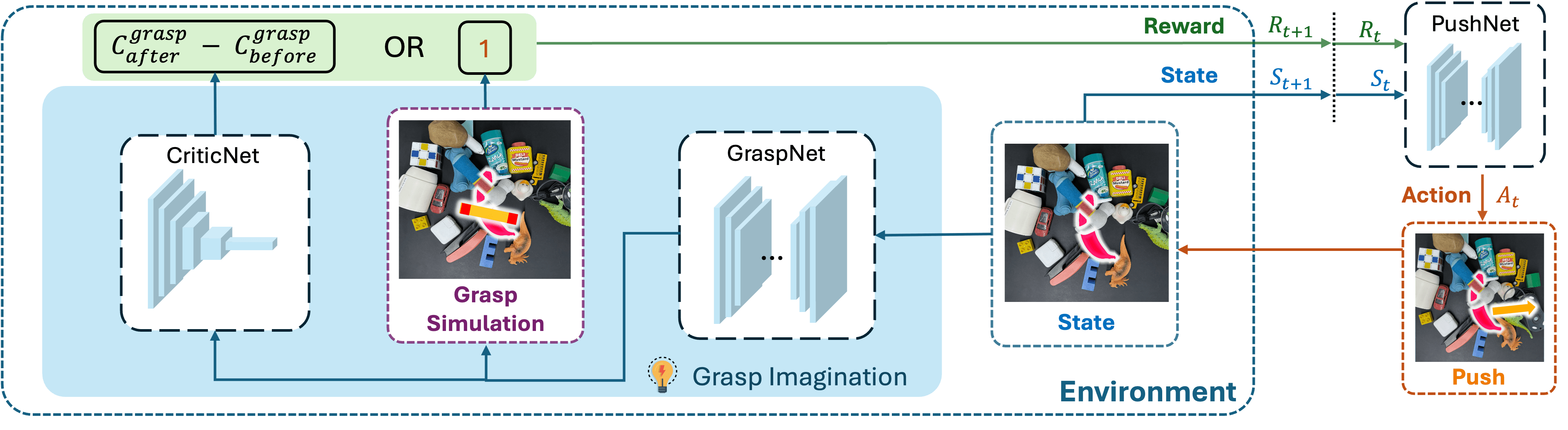}
\vspace{-0.2cm}
\caption{PushNet Training and CriticNet Finetuning Pipeline. The push reward is derived from the Grasp Imagination Module: it is 1 if the imagined grasp succeeds, otherwise it equals the difference in predicted grasp scores before and after the push.
}
\label{fig:train_pipeline}
\vspace{-0.5cm}
\end{figure*}

Similarly, CriticNet $\sigma$ is trained on the same dataset. However, unlike $\pi$, the model $\sigma$ receives $D_t$, the target object mask $k$, the object mask set $\mathcal{K}$, and a single grasp pose. Both $k$ and $\mathcal{K}$ are represented as binary maps, and the grasp pose is drawn on a blank image. All three maps have the same spatial size as $D_t$. The network then outputs a scalar score evaluating the grasp quality. Formally, $\sigma: \mathbb{R}^{4 \times h\times w} \to \mathbb{R}$. CriticNet is trained using the Mean Square Error (MSE) loss, defined as: 
\begin{equation}
\mathcal{L}_{\sigma} = \frac{1}{N} \sum_{i} (y - \hat{y} )^2
\vspace{-0.1cm}
\end{equation}
where $\hat{y}$ is the predicted grasp quality score, and $y$ is the ground-truth label corresponding to the given grasp pose. While both networks predict grasp scores, their roles differ: $\pi$ provides a pixel-wise quality estimation, whereas $\sigma$ measures a more accurate feasibility of a specific pose for the target.

\textbf{Step 2: PushNet Training and CriticNet Fine-tuning.}
This step is formulated as a \textbf{contextual bandit} learning problem, as shown in Figure \ref{fig:train_pipeline}. Unlike previous methods that perform complex alternative training schemes, we treat $\pi$ and $\sigma$ as part of the bandit environment to supervise the PushNet $\phi$ training. To enable this, we introduce the \textit{Grasp Imagination Module}, which provides a pushing reward by (1) simulating the optimal grasp predicted by $\pi$ in the post-push scene, and (2) evaluating the optimal grasp using $\sigma$. After evaluation, the simulation is restored to the post-push scene (i.e., before the grasp). As a result, the bandit environment is composed of two components: the cluttered physical scene itself and the Grasp Imagination Module. 

Specifically, after the simulation scene is initialized, the cluttered environment is segmented into multiple object masks using a segmentation model. The Grasp Imagination Module first stores the initial state and then simulates grasp attempts for each mask sequentially. After each simulated grasp, the environment is restored to the initial state. This process continues until a grasp failure occurs, at which point the training episode for pushing begins. PushNet $\phi$ will predict the $Q$ value for all pushing actions, and an $\epsilon$-greedy policy will be executed. After the push action, the Grasp Imagination Module simulates the grasp action again to assess the new grasp feasibility. If the grasp succeeds, the push action is considered optimal, and the reward is set to 1. If the grasp fails, we use an adaptive reward defined as the difference between the predicted grasp scores before and after the push estimated by $\sigma$, as a good push should improve the grasp feasibility. The episode terminates when a simulated grasp succeeds or the maximum pushing attempts are reached. The system then moves on to the next target mask or re-initializes the scene if all masks are iterated.

The PushNet $\phi$ takes $D_t$, the target object mask $k$, and the object mask set $\mathcal{K}$ as input and outputs pixel-wise push score maps with the same action space as $\pi$, i.e., $\phi: \mathbb{R}^{3\times h\times w} \to \mathbb{R}^{n\times h\times w}$. Its training objective is to minimize the following Huber loss: 
\begin{equation}
\mathcal{L}_{\phi} = \begin{cases}
\frac{1}{2}(r - Q_a)^2 & \text{if } |r - Q_a| \leq 1 \\
\delta \left( |r - Q_a| - \frac{1}{2} \right) & \text{otherwise}
\end{cases}
\vspace{-0.1cm}
\end{equation}
where $a$ and $r$ are the selected action and corresponding reward, and $Q_a$ represents the predicted push score for action $a$ produced by the PushNet $\phi$.

Meanwhile, CriticNet $\sigma$ is finetuned at this stage to align with the grasp distribution generated by a learned policy. Previously, $\sigma$ was trained using grasps from a random policy, but now it needs to evaluate grasps from $\pi$. To address the distribution shift, we use the grasp attempts predicted by $\pi$ in the Grasp Imagination Module, with their outcomes, to further optimize $\sigma$.

Our method has several advantages compared with \cite{xu2021efficient,wang2022self}. First, generating rewards from the network’s predictions enables self-supervised learning, eliminating the need for manual evaluation of push effectiveness. Second, our method is highly flexible, since $\phi$ and $\sigma$ can be trained to align with any grasp policy. This enables adaptation to different grasp strategies and enhances robustness. In the experiments, we demonstrate that our framework can also improve the performance of other trained grasp networks.

\begin{figure}[!t]
    \centering
    \includegraphics[width=0.9\linewidth]{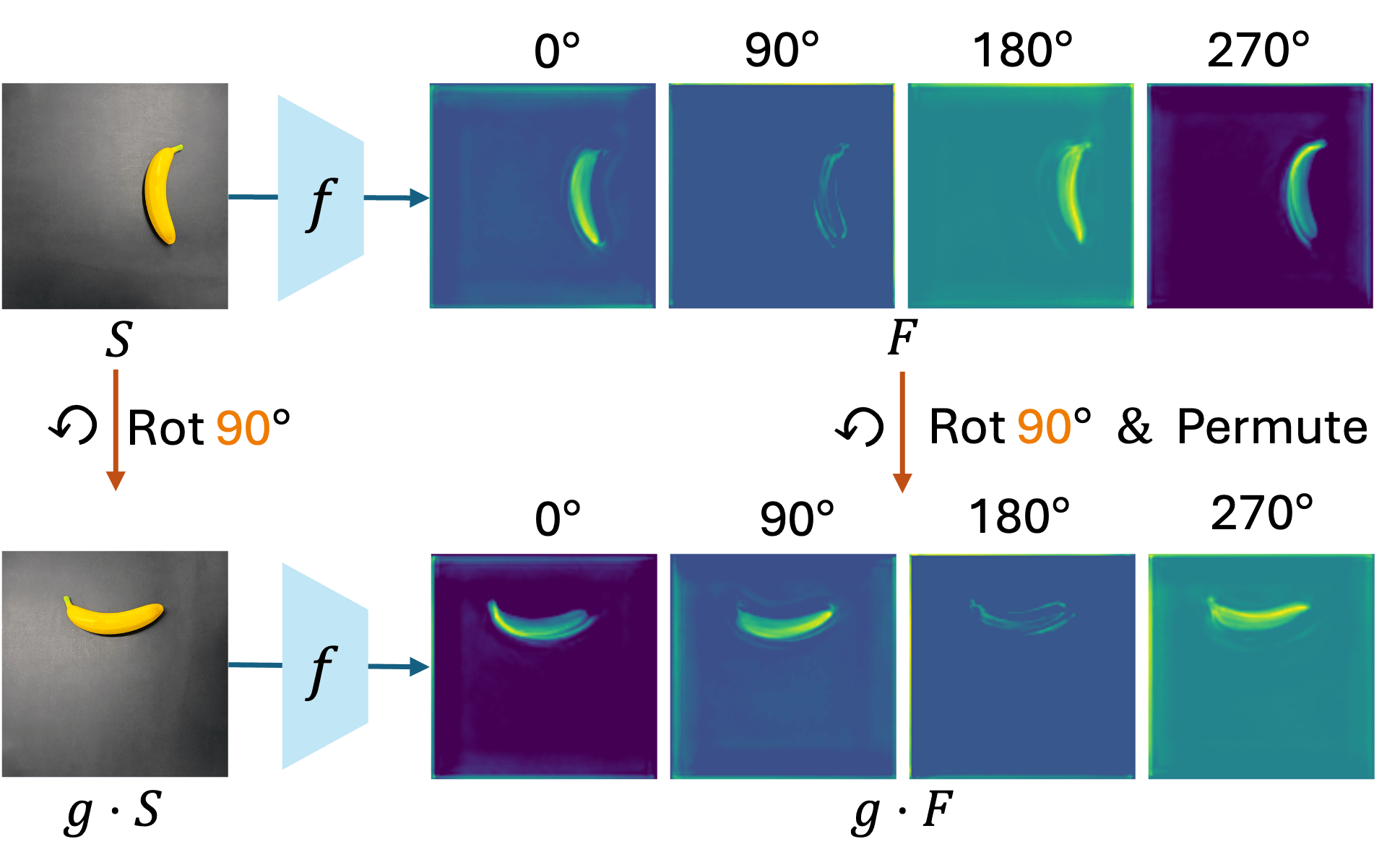}
    \vspace{-0.2cm}
    \caption{Illustration of how an element $g$ acts on feature maps by rotating the pixels and permuting the order of the channels. The angles above feature maps indicate the candidate grasp orientations.
    }
    \label{fig:EbquiRepr}
    \vspace{-0.5cm}
\end{figure}

\subsection{Equivariance and Invariance in Agent Learning}
A network $h$ is equivariant to a symmetry group $G$ if for all $g \in G$, it satisfies: $h(g \cdot x) = g \cdot h(x)$.
This property ensures that applying a transformation $g$ to the input results in an equivalent transformation in the output. In particular, if the symmetry group is $G =\SE(2)$ (i.e., rotation around the z-axis of the world frame and translation along the x and y-axes), a planar rotation and translation of the input results in the same rotation and translation to the output. This symmetry naturally reflects the inherent structure of many table-top robotic tasks, such as grasping and pushing, while avoiding learning unnecessary out-of-plane rotation equivariance (e.g., full $\SO(3)$ rotations), which is both redundant and computationally expensive.

Specifically, 
we design GraspNet $\pi$ and PushNet $\phi$ to be \textbf{equivariant} under the product group $C_n \times \mathbb{T}^2$, where $C_n=\{2\pi m/n:0\leq m < n\} \subset \SO(2)$, with $n\in \mathbb{Z}$, is a finite cyclic group of planar rotations, and $\mathbb{T}^2$ represents 2D translations. For either network $f \in \{\pi, \phi\}$, the equivariance property holds:
$
f(g \cdot \mathcal{I}) = g \cdot f(\mathcal{I}), \forall g \in C_n \times \mathbb{T}^2
$,
where $\mathcal{I}$ denotes the network input, which differs for $\pi$ and $\phi$. The group action $g\in C_n$ transforms the output map of shape $n\times h\times w$ by rotating the spatial dimensions $h\times w$ and cyclically permuting the orientation channels indexed by $n$, as illustrated in Figure~\ref{fig:EbquiRepr}. For further details on the rotational equivariance, see~\cite{wang2022equivariant, wang2022mathrm, zhu2022sample}. CriticNet $\sigma$ is designed to be \textbf{invariant} to the same group transformation in $\SE(2)$. This is a special case of equivariance where the output remains unchanged when the input $\mathcal{I}$ is transformed, since transforming $\mathcal{I}$ corresponds to transforming both the observation and the grasp action simultaneously, i.e., $
\sigma(g \cdot \mathcal{I}) = \sigma(\mathcal{I})$.
\subsection{Network Architectures}
To achieve the desired equivariance properties, we inherently achieve translational equivariance through Fully Convolutional Networks~\cite{long2015fully} and explicitly implement rotational ($\SO(2)$) equivariance using the \texttt{escnn} library~\cite{cesa2022program}. Separate architectures are designed for grasping and pushing policies to capture task-specific features.

In particular, GraspNet $\pi$ and CriticNet $\sigma$ are designed to predict and evaluate grasp poses, relying primarily on the accurate perception of local geometric structures. To achieve this, we adopt a ResNet~\cite{he2016deep} architecture for $\sigma$ and a U-Net architecture for $\pi$, both equivariant under the cyclic group $C_6$. A group pooling layer at the end of $\sigma$ transforms its representation from equivariant to invariant. To accurately predict grasp orientations, we introduce a finer-grained orientation representation within the $C_6$ framework of $\pi$. Specifically, we define each group element of $C_6$ to act on six orientation sub-channels, collectively forming a 36-dimensional orientation representation with intervals of $10^\circ$. The group action cyclically permutes these sub-channels in increments of $60^\circ$. Furthermore, the bilateral symmetry of the gripper implies that a rotation of $180^\circ$ about its forward axis leaves the grasp outcome unchanged, reducing the prediction range of grasp orientation from $360^\circ$ to $180^\circ$. This symmetry simplification can be mathematically formalized as a quotient representation of $\SO(2)$ w.r.t the $C_2$ subgroup, which identifies orientations that differ by $180^\circ$, as discussed in~\cite{zhu2022sample}. As a result, GraspNet $\pi$ achieves an angular resolution of $10^\circ$ over a $180^\circ$ rotation range while preserving $C_6$ equivariance.

\begin{figure}[!t]
    \centering
    \includegraphics[width=\linewidth]{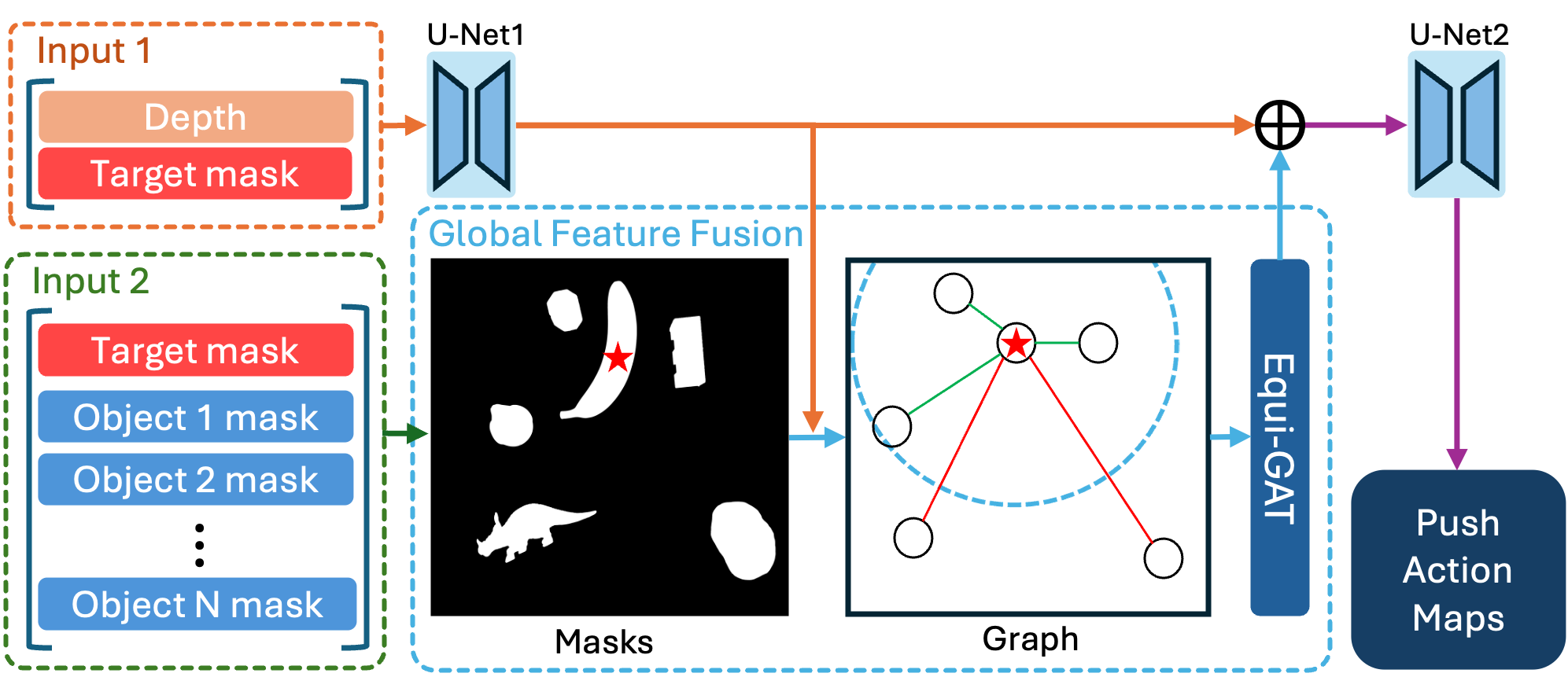}
    \caption{\textbf{PushNet Structure}.
   In the graph, the target node (red star) connects to nearby nodes within a predefined distance threshold (blue circle). Green edges are valid connections, while red edges are invalid.
}
    \label{fig:PushNet}
    \vspace{-0.6cm}
\end{figure}

In contrast to $\sigma$ and $\pi$, PushNet $\phi$ requires both global geometric context of the scene and local features of surrounding objects. As shown in Figure~\ref{fig:PushNet}, $\phi$ first extracts global features through an equivariant U-Net. To integrate local context, we introduce a feature fusion block. Here, feature maps from the U-Net are segmented by object masks, with each masked region serving as a node in a graph. Edges between nodes are formed based on the spatial distance between the target and surrounding objects. An Equivariant Graph Attention Layer then processes this graph to capture object interactions. The enriched graph features are merged with the original U-Net features and further refined by a second equivariant U-Net, yielding the final Q-value map for push action selection. Similar to $\pi$, $\phi$ employs three orientation sub-channels for each $C_6$ group element. However, unlike grasping, pushing requires full $360^\circ$ rotational coverage due to its directional nature, which breaks $180^\circ$ rotational invariance. Consequently, $\phi$ achieves $20^\circ$ orientation resolution across the full $360^\circ$ rotation range.

%%%%%%%%%%%%%%%%%%%%%%%%%%%%%%%%%%%%%%%%%%%%%%%%%%%%%%%%%%%%%%%%%%%%%%%%%%%%%%%%

\section{Experiments}

\subsection{Training Details}
In step 1, we randomly initialize scenes in simulation with 2-15 objects and use SAM2 to generate the mask set. For each mask, we randomly sample 600 grasp poses and record the grasp outcomes. In total, we collect 3.6M grasp data points (approximately 2M positive and 1.5M negative). GraspNet is trained for 30 epochs, while CriticNet is trained for 15 epochs. In step 2, PushNet is trained for 2,000 steps, with CriticNet fine-tuned for the same number of steps. The push action is parameterized as a 10 cm movement along the target direction. All networks are trained on simulation data and directly transferred to real-world settings.

\subsection{Experienment Setups and Tasks}

\begin{table*}[!t]
    \renewcommand\arraystretch{0.8}
    \setlength{\tabcolsep}{8.3pt}
    \centering
    \caption{\textbf{Goal-conditioned Push-Grasp in Clutter} in simulation. All methods are allowed a maximum of 5 push attempts per target object. For each object count setting, testing is conducted using 4 random seeds, each with $n$ rounds, where $n$ is set to 300 divided by the object count (e.g., $n = 30$ for 10 objects). At the beginning of each round, SAM2 generates masks for all objects in the initial scene. Then, each mask, with its corresponding object, is sequentially selected as the target from the initial scene state. SAM2 is continuously used to track the target and other object masks during pushing, and a grasp is attempted if the grasp score is above the threshold or the limit of 5 pushes is reached. After each grasp attempt, the environment is reset to its initial state before proceeding to the next target. Final results are averaged over the 4 seeds.}
    \vspace{-0.1cm}
    \begin{tabular}{l c c c c c c c c}
        \toprule
        Method & \multicolumn{2}{c}{10 Objects} & \multicolumn{2}{c}{15 Objects} & \multicolumn{2}{c}{20 Objects} & \multicolumn{2}{c}{25 Objects}\\
        
         & GSR (\%) & ME (\%) & GSR (\%) & ME (\%) & GSR (\%) & ME (\%) & GSR (\%) & ME (\%) \\
        \midrule
        \citet{xu2021efficient} &43.5&49.9&33.9&41.7&25.6&40.8&24.6&42.5\\
        \citet{wang2022self} &54.7&42.9&47.9&28.5&42.2&27.9&39.2&27.7\\
        \cite{wang2022self} (Grasp) + Ours (Push) &56.1&54.5&50.2&55.1&55.2&33.0&49.5&34.7\\
        \cite{wang2022self} (Push) + Ours (Grasp) &90.6&72.9&89.0&\textbf{77.1}&87.5&\textbf{79.4}&82.4&\textbf{81.3}\\
        Ours (non-equi + data aug)
        &81.2&49.8&82.5&49.8&81.0&56.3&74.6&55.9\\
        \textbf{Ours} &\textbf{97.0}&\textbf{77.6}&\textbf{95.1}&69.0&\textbf{95.0}&65.2&\textbf{92.0}&57.2 \\
        \hline
    \end{tabular}

    \label{tab:task1}
    % \vspace{-0.2cm}
\end{table*}

\begin{table*}[!ht]
    \renewcommand\arraystretch{0.8}
    \setlength{\tabcolsep}{9.5pt}
    %\captionsetup{font=footnotesize}
    \centering
    \caption{\textbf{Clutter Clearing} in simulation. Each result is reported in a \textit{without} / \textit{with} push action format to evaluate the effectiveness of pushing. The evaluation follows the Goal-conditioned Push-Grasp in Clutter protocol, where 4 seeds are used, each running $n$ rounds, with $n=300/\text{object counts}$. A maximum of 5 push attempts is allowed. Final results are averaged over the 4 seeds.}
    \vspace{-0.1cm}
    % \scriptsize
    \begin{tabular}{l c c c c c c c c}
        \toprule
        Method &  \multicolumn{2}{c}{10 Objects} & \multicolumn{2}{c}{15 Objects} & \multicolumn{2}{c}{20 Objects} & \multicolumn{2}{c}{25 Objects}\\
        
         & GSR (\%) & DR (\%) & GSR (\%) & DR (\%) & GSR (\%) & DR (\%) & GSR (\%) & DR (\%) \\
        \midrule
        \citet{xu2021efficient}&60.8/60.6&40.7/40.8&57.5/56.2&29.3/27.2&55.2/51.1&18.9/17.9&51.0/51.4&13.8/14.6  \\
        
        \citet{wang2022self}&56.0/54.1&38.9/35.2&59.5/59.3&31.6/31.0&59.1/57.6&23.7/22.5&54.6/60.2&16.0/23.8\\
        
        Ours&\textbf{83.0}/\textbf{97.5}&\textbf{69.2}/\textbf{93.9}&\textbf{83.3}/\textbf{97.6}&\textbf{62.0}/\textbf{91.1}&\textbf{83.6}/\textbf{97.7}&\textbf{53.2}/\textbf{91.1}&\textbf{80.4}/\textbf{97.8}&\textbf{35.5}/\textbf{87.8} \\
        \hline

    \end{tabular}
    %\vspace{-0.1cm}

    \label{tab:declutter}
    \vspace{-0.5cm}
\end{table*}

\begin{figure}[t]
    \centering
    % \captionsetup{font=footnotesize}
    % \scriptsize
    \includegraphics[width=0.65\linewidth]{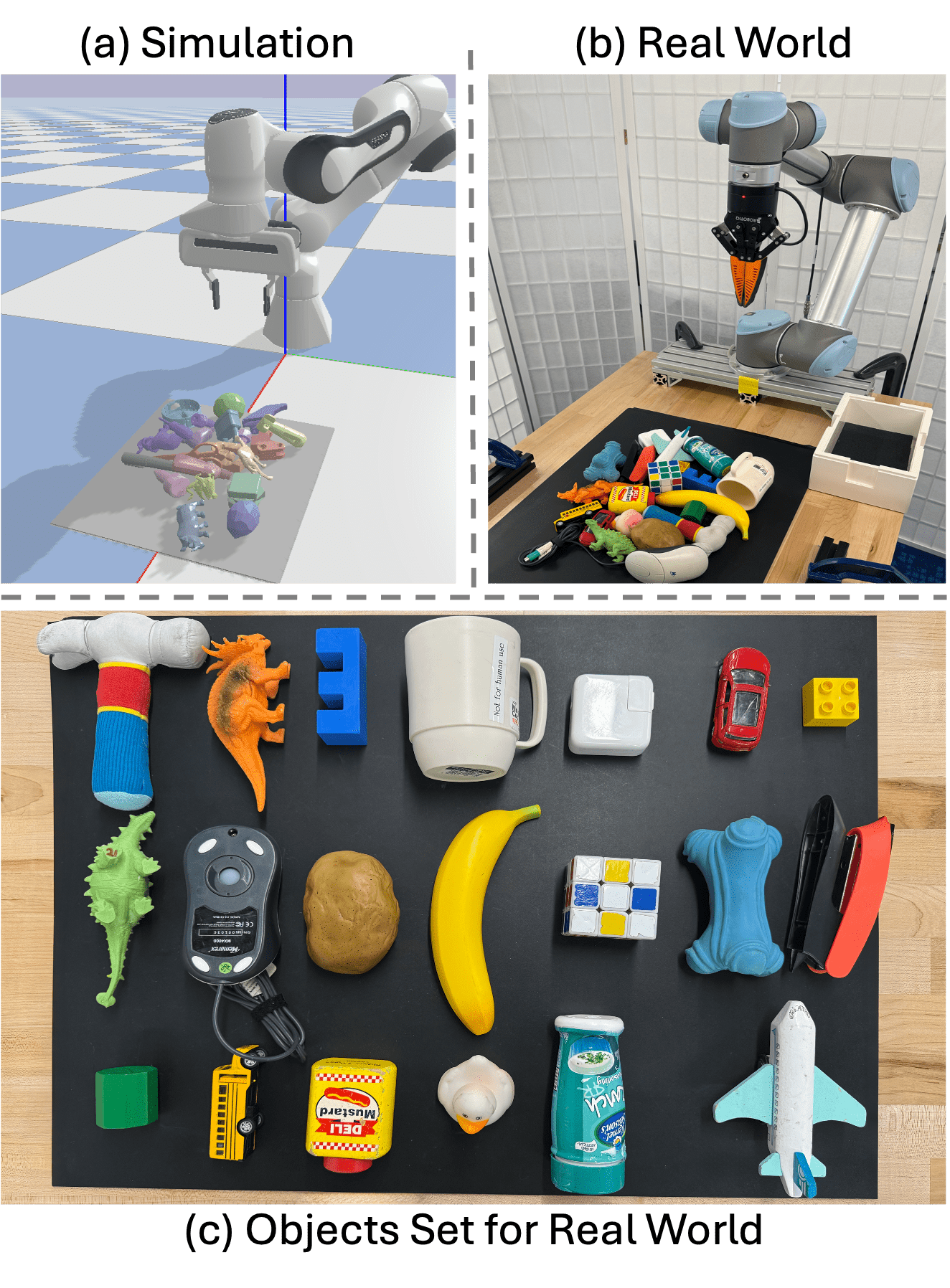}
    \vspace{-0.2cm}
    \caption{\textbf{Experiment Setup.} The workspace is a 40 $cm^3$ cube in both environments. The training and test object sets in simulation follow~\cite{zhu2022sample}, while the real-world object set is shown in (c).}
    \label{fig:experienment_setup}
    \vspace{-0.7cm}
\end{figure}

\begin{figure*}[!h]
    \captionsetup[subfigure]{labelformat=empty}
    %$\captionsetup{font=footnotesize}
    \centering
    \scriptsize
    \begin{subfigure}[t]{0.12\linewidth}
        \centering
        \includegraphics[width=1\linewidth]{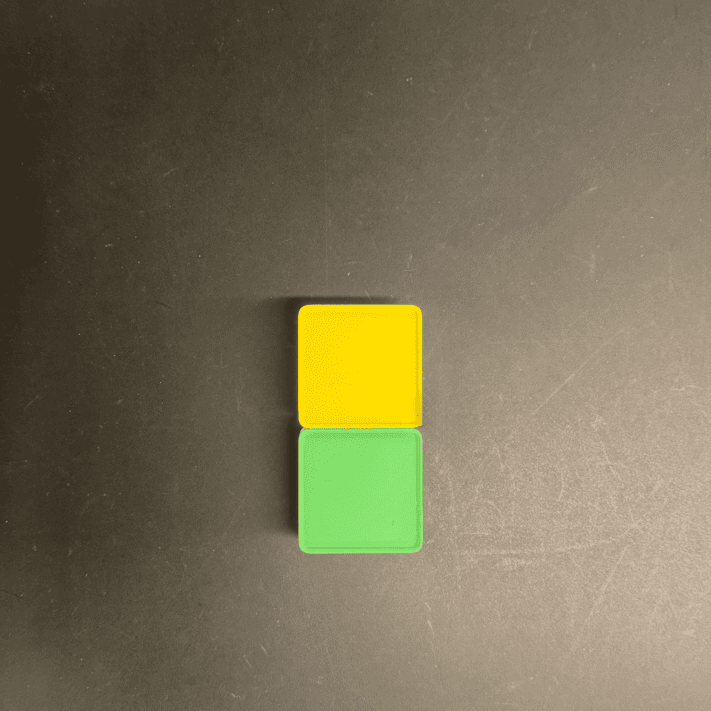}
        \vspace{-0.4cm}
        \caption{Case 1}
        
    \end{subfigure}
    \begin{subfigure}[t]{0.12\linewidth}
        \centering
        \includegraphics[width=1\linewidth]{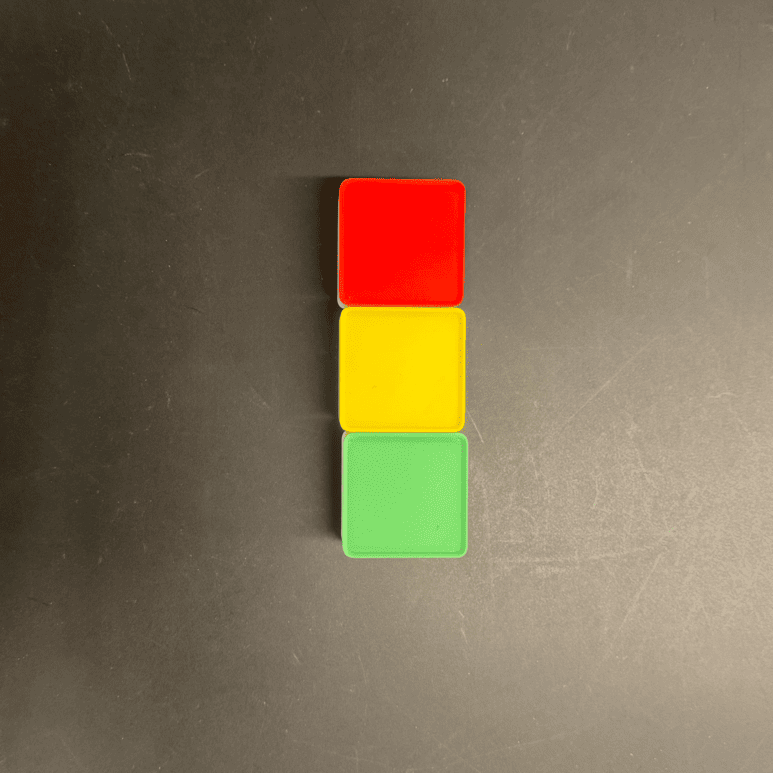}
        \vspace{-0.4cm}
        \caption{Case 2}
        
    \end{subfigure}
    \begin{subfigure}[t]{0.12\linewidth}
        \centering
        \includegraphics[width=1\linewidth]{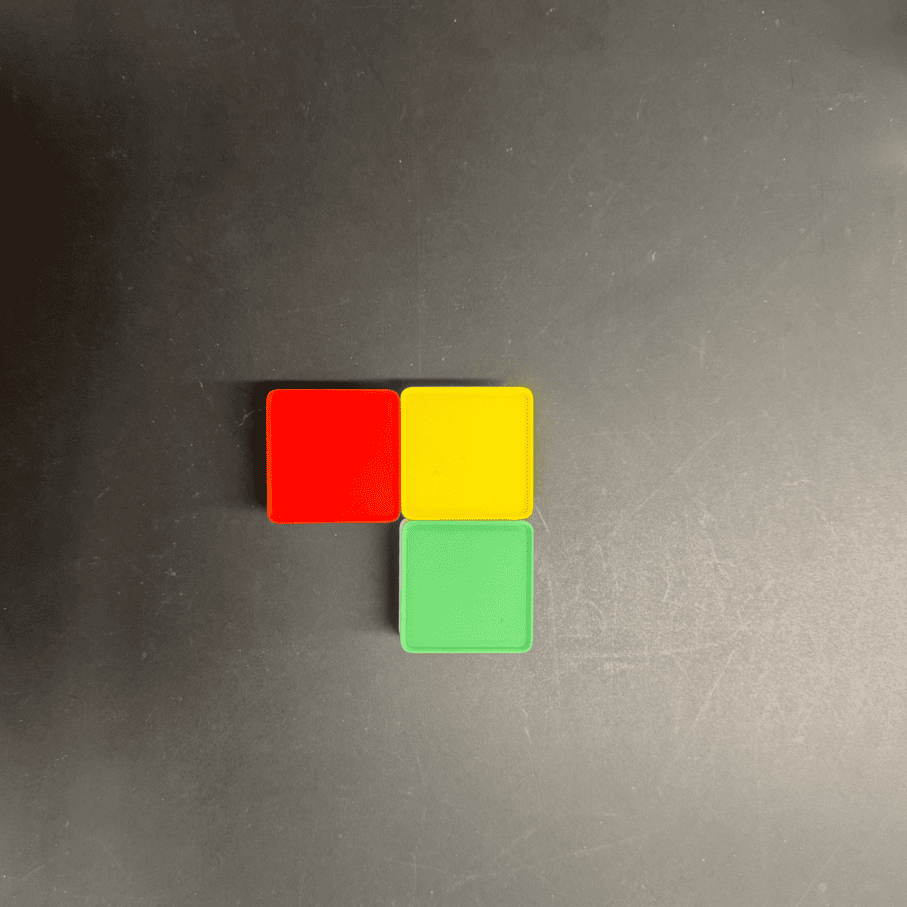}
        \vspace{-0.4cm}
        \caption{Case 3}
        
    \end{subfigure}
    \begin{subfigure}[t]{0.12\linewidth}
        \centering
        \includegraphics[width=1\linewidth]{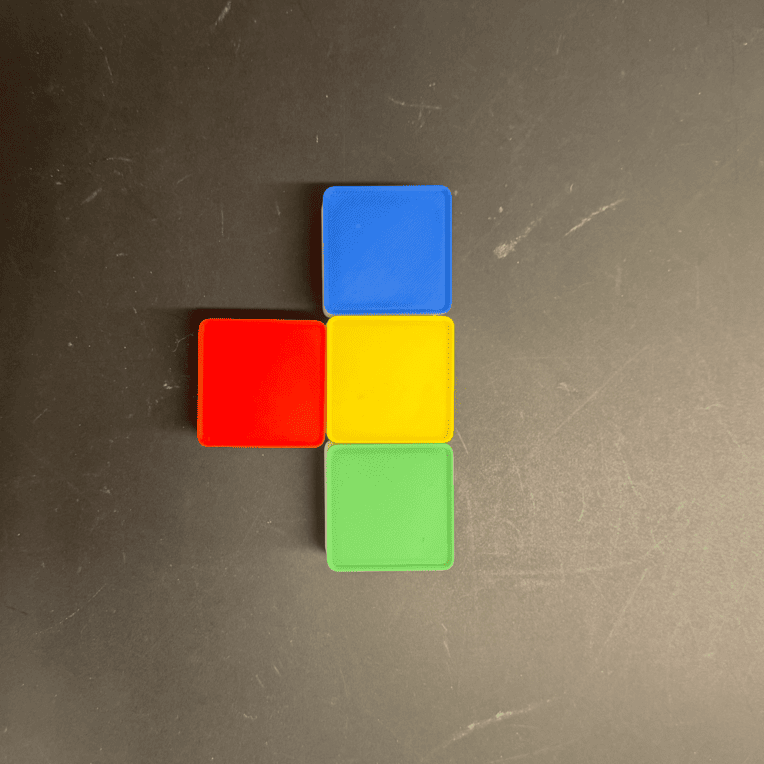}
        \vspace{-0.4cm}
        \caption{Case 4}
        
    \end{subfigure}
    \begin{subfigure}[t]{0.12\linewidth}
        \centering
        \includegraphics[width=1\linewidth]{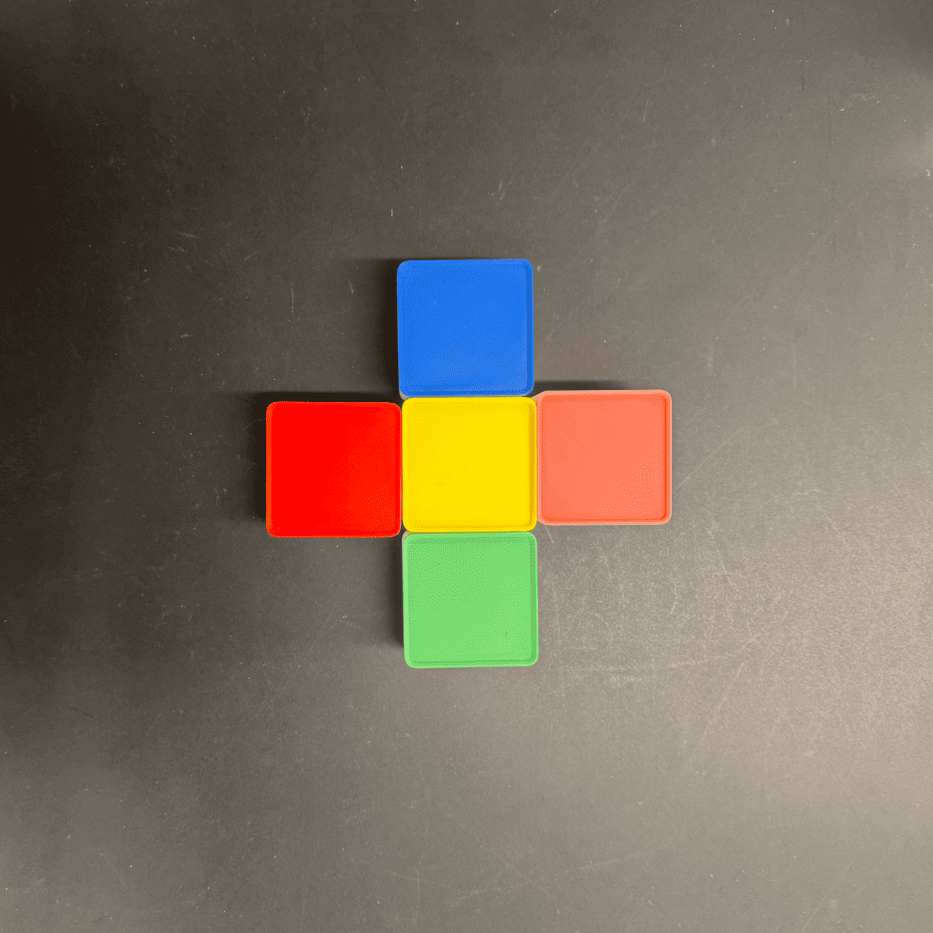}
        \vspace{-0.4cm}
        \caption{Case 5}
        
    \end{subfigure}
    \begin{subfigure}[t]{0.12\linewidth}
        \centering
        \includegraphics[width=1\linewidth]{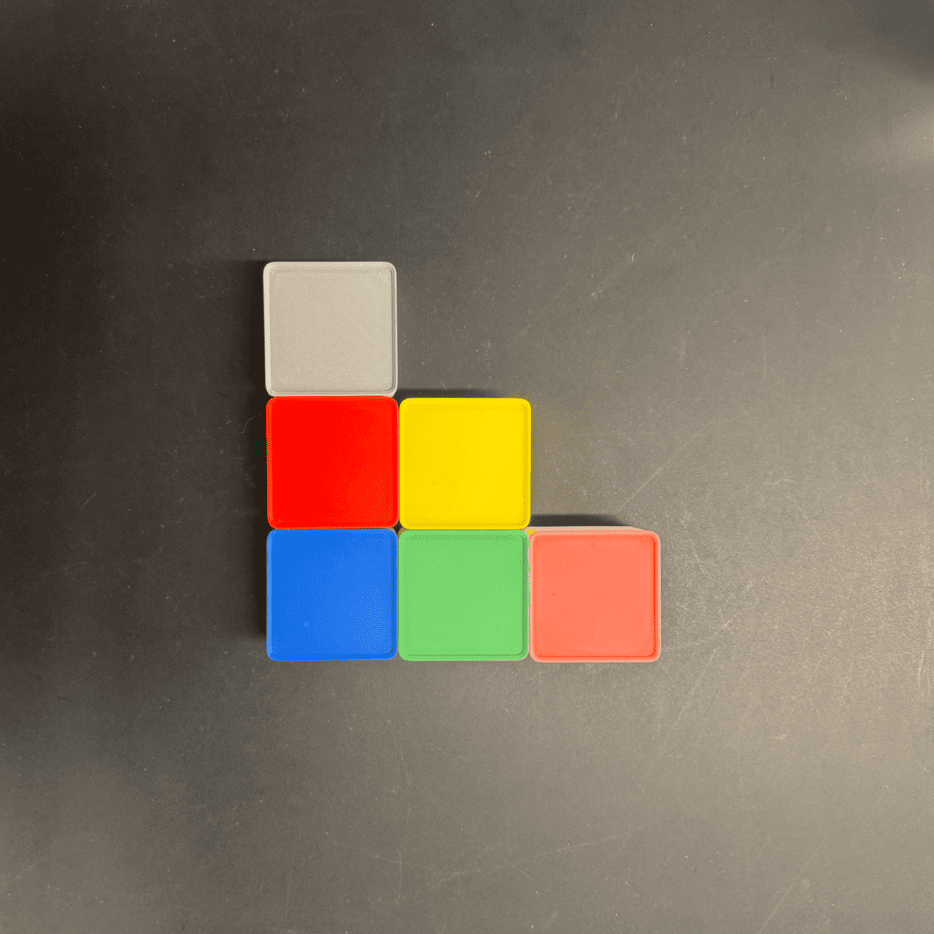}
        \vspace{-0.4cm}
        \caption{Case 6}
        
    \end{subfigure}
    \begin{subfigure}[t]{0.12\linewidth}
        \centering
        \includegraphics[width=1\linewidth]{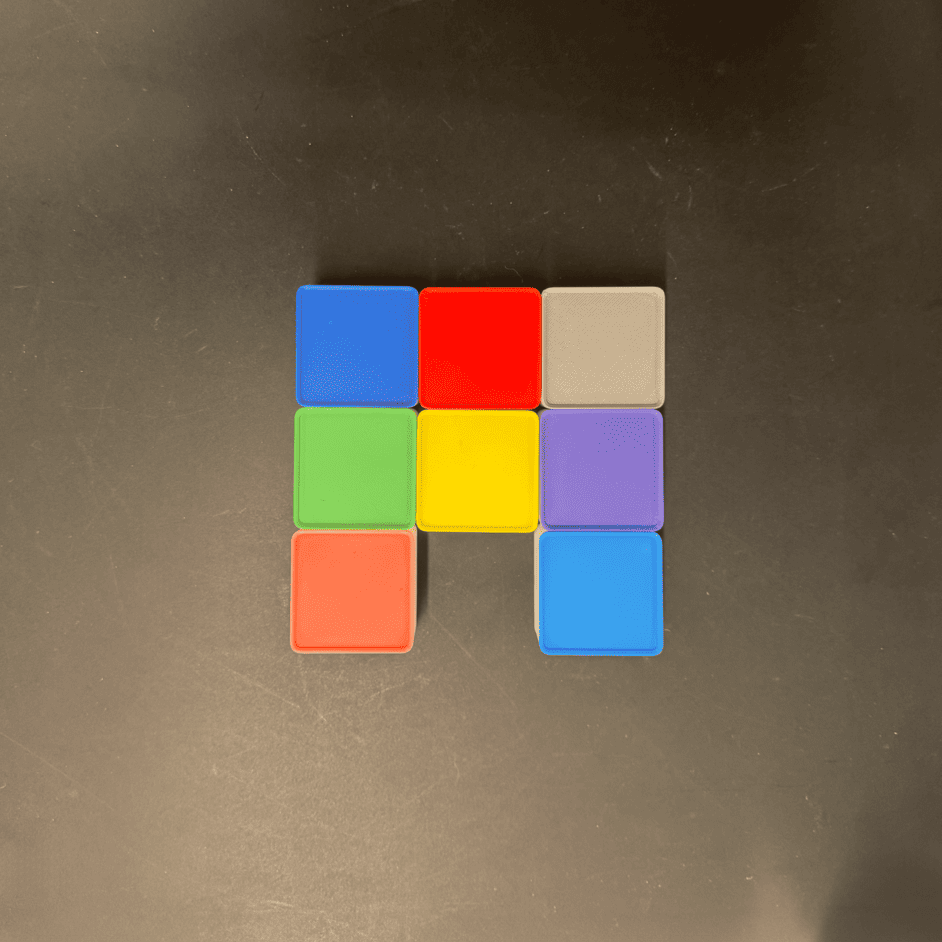}
        \vspace{-0.4cm}
        \caption{Case 7}
        
    \end{subfigure}
    \begin{subfigure}[t]{0.12\linewidth}
        \centering
        \includegraphics[width=1\linewidth]{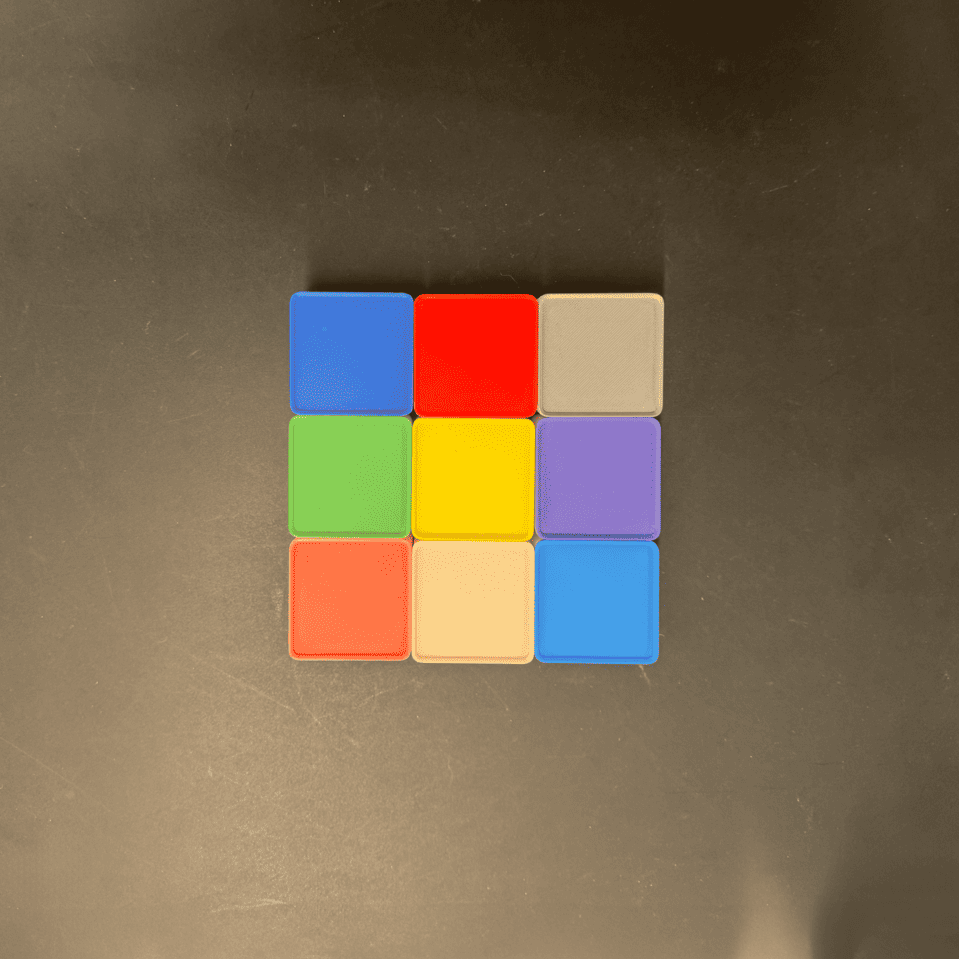}
        \vspace{-0.4cm}
        \caption{Case 8}
        
    \end{subfigure}
    \vspace{-0.1cm}
    \caption{Eight special configurations of the \textbf{Goal-conditioned Push-Grasp in Constrained Spaces} task in the real world}.
    \label{fig:specialcase_real}
    \vspace{-0.3cm}
\end{figure*}

To evaluate our push-grasp framework, we conduct experiments in both simulation and real-world environments, with the setup illustrated in Figure \ref{fig:experienment_setup}. The evaluation consists of three tasks:\\
\textbf{Goal-Conditioned Push-Grasp in Clutter.} This task assesses our framework's ability to retrieve a specific object from a cluttered scene. Following \cite{huang2022edgegraspnetworkgraphbased, hu2024orbitgrasp}, objects are randomly initialized, but with one designated as the target. The robot performs push actions if needed before grasping the target.\\ 
\textbf{Clutter Clearing.} This task evaluates the ability to clear an entire scene without any predefined target or grasp sequence. The setup follows the previous task.\\
\textbf{Goal-Conditioned Push-Grasp in Constrained Spaces.} Figure \ref{fig:specialcase_real} shows the task configuration. Objects are arranged in challenging geometric configurations (e.g., tight clusters, narrow gaps). This is a hard task because the robot must push strategically to create graspable space in a constrained environment.
\subsection{Evaluation Metrics and Baselines}
We use three evaluation metrics:
\textbf{Grasp Success Rate (GSR)}, the ratio of successful grasps to total grasp attempts; \textbf{Declutter Rate (DR)}, the proportion of grasped objects relative to the total number of objects; and \textbf{Motion Efficiency (ME)}~\cite{zeng2018learning}, the fraction of grasp actions among all executed actions. GSR is used for all tasks, with DR applied to clutter-clearing and ME to goal-conditioned tasks.
Our method are compared with two baselines:
(1) \textbf{\citet{xu2021efficient}}, a goal-conditioned push-grasp framework that utilizes multi-stage training to jointly optimize push and grasp action prediction.
(2) \textbf{\citet{wang2022self}}, an extension of \cite{xu2021efficient} that improves performance by relaxing the constraints on Q-value selection and using object masks to guide actions. In addition, we introduce three ablation variants to highlight our framework design. The first integrates the grasp module from \cite{wang2022self} into our framework, while the second applies our GraspNet within the framework of \cite{wang2022self}. The third replaces the equivariant network with non-equivariant counterparts, trained with data augmentation.
\begin{table*}[!t]
\renewcommand\arraystretch{0.8}
    \caption{Real-world \textbf{Goal-conditioned Push-Grasp in Constrained Spaces} results, reported as “successful iterations / total iterations”.}
    \setlength{\tabcolsep}{12.5pt}
    %\captionsetup{font=footnotesize}
    \centering
    % \scriptsize
    \vspace{-0.1cm}
    \begin{tabular}{l c c c c c c c c}
        \toprule
         Method & Case 1 & Case 2 & Case 3 & Case 4 & Case 5 & Case 6 & Case 7 & Case 8 \\
        \midrule
        \citet{xu2021efficient} & 9/10 & 4/10 & 4/10 & 5/10 & 4/10 & 3/10 & 4/10 & 4/10 \\
        \citet{wang2022self} & 6/10 & 7/10 & 6/10 & 5/10 & 3/10 & 4/10 & 4/10 & 5/10 \\
        Ours & \textbf{10/10} & \textbf{10/10} & \textbf{9/10} & \textbf{8/10} & \textbf{7/10} & \textbf{10/10} & \textbf{9/10} & \textbf{8/10} \\
        \hline
        
    \end{tabular}
    \label{tab:specialcase_real}
    \vspace{-0.5cm}
\end{table*}
\subsection{Comparison with Baseline Methods in Simulation}
We report the comparison result for the \textbf{Goal-conditioned Push-Grasp in Clutter} task in Table \ref{tab:task1}. Our method achieves the best performance, significantly outperforming all baselines. On average, across all the settings with different number of objects, it surpasses the best baseline~\cite{wang2022self} by \textbf{48.8\%} in GSR. The first two variations (Table \ref{tab:task1}, row 3 and 4) show that integrating our approach into existing baselines further improves their performance, which highlights our design's effectiveness. However, our PushNet within the framework of ~\cite{wang2022self} does not yield significant improvement over the original method. This is likely because, while PushNet successfully creates graspable space, the baseline grasp module lacks sufficient capability to retrieve targets. The third variant serves two purposes: it first proves the advantage of equivariant networks over non-equivariant counterparts with data augmentation, and it further validates the effectiveness of our train pipeline compared to baseline training strategies. Although our method's ME is similar to baselines, this is expected, as additional push actions are necessary to ensure more successful grasps.

\begin{figure}[t!]
    %\captionsetup{font=footnotesize}
    \centering
    \scriptsize
\includegraphics[width=0.9\linewidth]{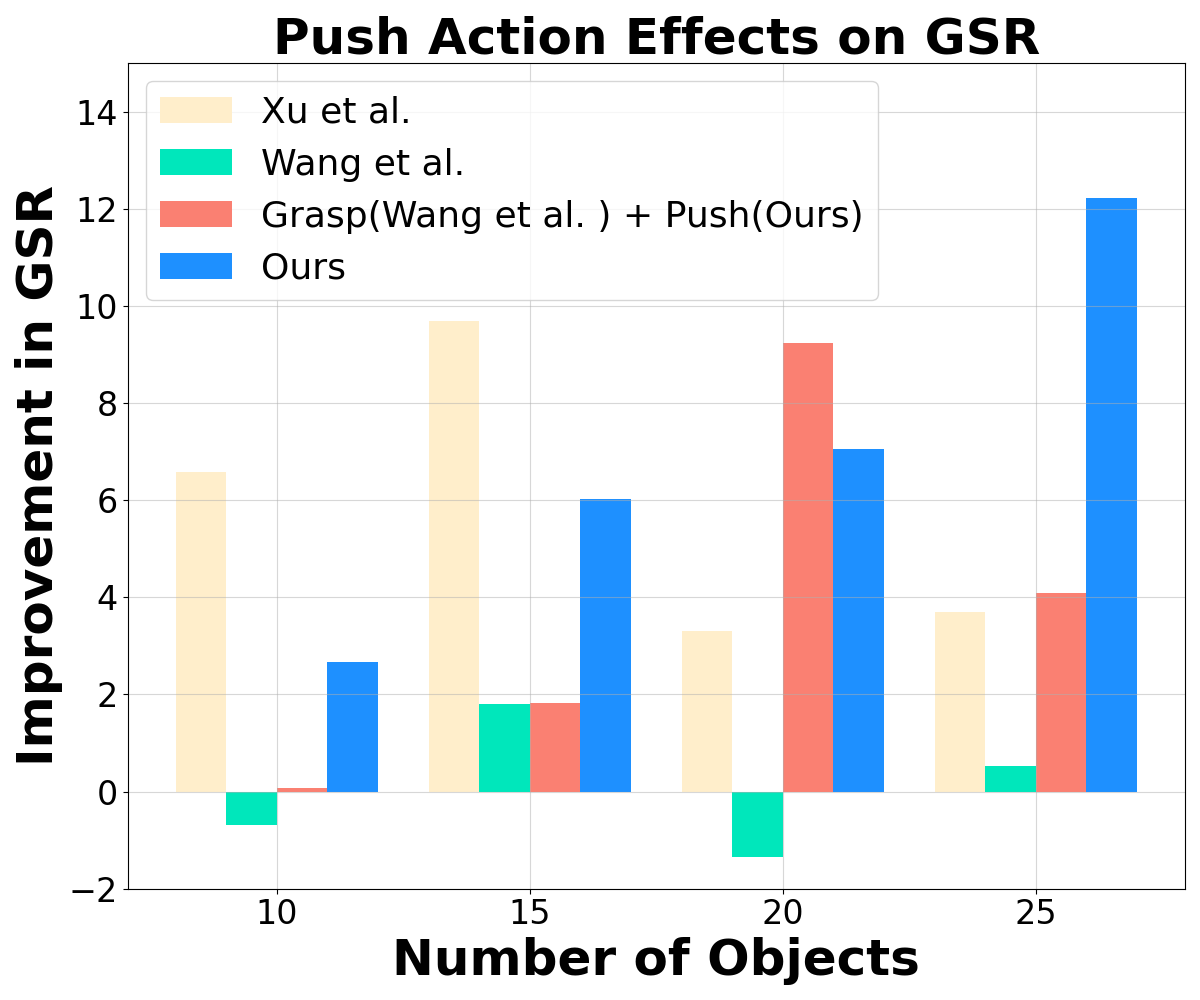}
    \vspace{-0.2cm}
    \caption{Improvements in GSR with and without push actions, measured as the difference between 5 pushes and 0 pushes.}
    \label{fig:diff}
    \vspace{-0.5cm}
\end{figure}

We also conduct an ablation study for this task, as shown in Figure \ref{fig:diff}. The bar chart compares the improvement in GSR with and without push actions. Our PushNet improves GSR by approximately 12\%  in highly cluttered environments. Additionally, we observe that the push module in ~\citet{wang2022self} contributes little to improving GSR, whereas integrating our PushNet leads to a more significant improvement.

Table \ref{tab:declutter} shows the results of the \textbf{Clutter Clearing} task. Although this task is target-agnostic, push actions remain beneficial in cluttered environments. Since there is no specific target, we use the object with the highest score from GraspNet as the target object for each step. The results show that our method's grasping capability exceeds all baselines by a large margin in both with and without push actions. Notably, even without push actions, our method consistently outperforms all baselines that employ pushing, across all settings with different numbers of objects. This highlights the strong capacity of our GraspNet to handle cluttered scenes. Furthermore, when push actions are enabled, our method achieves additional improvements. The magnitude of this improvement is significantly greater than that observed in any of the baselines, demonstrating the strong contribution of our PushNet in creating graspable space.

\subsection{Real World Experiments}
\begin{table}[t!]
\renewcommand\arraystretch{0.8}
    \vspace{0.2cm}
    \caption{Real world \textbf{Goal-conditioned Push-Grasp in Clutter} comparison results. GSR is reported as “successful grasps / total attempts”, while ME is defined as "total grasp attempts / total actions". }
    \setlength{\tabcolsep}{12pt}
    %\captionsetup{font=footnotesize}
    \centering
    % \scriptsize
    \vspace{-0.1cm}
    \begin{tabular}{l c c c }
        \toprule
        Method & GSR(\%) & ME(\%)  \\
        \midrule
        \citet{xu2021efficient} &40 (40/100)&27.3 (100/367)\\
        \citet{wang2022self} &51 (51/100)& 26.8 (100/373) \\
        % Grasp(Wang et al.) + Push(Ours) & 0 (0/100) & 0.0 \\
        \textbf{Ours} & \textbf{86 (86/100)} & \textbf{38.6 (100/259)} \\
        \hline
    \end{tabular}
    \vspace{-0.2cm}
    \label{tab:realworld}
    \vspace{-0.3cm}
\end{table}
We conduct a large-scale real-world evaluation that far exceeds the number of trials in prior baseline studies. This extensive setup reduces the influence of randomness and increases the reliability of our results. To assess the performance of our method, we evaluate it on two tasks: \textbf{Goal-conditioned Push-Grasp in Clutter} and \textbf{Goal-conditioned Push-Grasp in Constrained Spaces}. The trained model is directly transferred from simulation to the real-world environment without any fine-tuning.

The \textbf{Goal-conditioned Push-Grasp in Clutter} task involves grasping randomly selected targets from a set of 20 household objects placed randomly in the workspace. The real-world setup and object sets are shown in Figure \ref{fig:experienment_setup}(b) and (c). The experimental protocol follows the simulation setup. Each run consists of attempting to retrieve five target objects, with the scene randomly rearranged after each grasp to create a new cluttered layout for the next target. Each method is evaluated over 20 runs (i.e., 100 target objects in total). The target object's mask is still tracked via SAM2. Table \ref{tab:realworld} presents the results, comparing our method with several baselines. Our EPG significantly outperforms all baselines by at least \textbf{35\%} in GSR. The primary failure cases are: 1) inaccurate object masks from SAM2, which further affect PushNet and CriticNet outputs; 2) imprecise grasp poses predicted by the GraspNet. Despite these challenges, our method demonstrates strong overall stability.

The configuration of the \textbf{Goal-conditioned Push-Grasp in Constrained Spaces} task is illustrated in Figure~\ref{fig:specialcase_real}. It contains eight different cases, each with a varying number of small boxes placed in specific positions. The objective is to grasp the \textbf{yellow box}, which is consistently placed at the center of surrounding boxes. These task configurations are \textbf{unseen during training} and require effective strategies to solve, placing a strong demand on the model’s generalization ability. For each case, experiments are conducted over 10 iterations, where each iteration involves a randomized scene rotation and a different arrangement of the boxes. The results in Table \ref{tab:specialcase_real} indicate that despite increasing task complexity, our method consistently outperforms the baselines while maintaining stable performance.

%%%%%%%%%%%%%%%%%%%%%%%%%%%%%%%%%%%%%%%%%%%%%%%%%%%%%%%%%%%%%%%%%%%%%%%%%%%%%%%%

\section{Conclusion and Limitation}
This paper introduces the \textbf{Equivariant Push-Grasp (EPG)} Network, a goal-conditioned grasping method that incorporates push actions to improve performance. EPG leverages $\SE(2)$-equivariance to enhance sample efficiency and generalization. We also propose a flexible training framework that optimizes PushNet using grasp score differences as rewards, avoiding manually designed reward functions and complex alternating training. Extensive experiments show that EPG consistently outperforms strong baselines across various tasks and settings. 

However, our method has several limitations. First, it operates in an open-loop manner with a fixed push distance, lacking real-time feedback for adaptive push strategies. Future work will explore closed-loop control. Second, EPG is limited to 4-DoF, which is sufficient for tabletop settings but does not generalize well to more complex 6-DoF scenarios. Extending to 6-DoF would enable broader applicability. Finally, EPG may require manual selection of target masks for consistency, which is inconvenient. To address this, we aim to integrate vision-language models for automatic mask generation.

% \bibliographystyle{IEEEtran}
% \bibliography{ref}
\bibliographystyle{abbrvnat}
\bibliography{ref}

% \printbibliography %added
\end{document}